\documentclass[final,5p,times,twocolumn,authoryear]{elsarticle}
\usepackage{amsmath,amssymb}
\usepackage{graphicx}
\usepackage{booktabs}
\usepackage[hidelinks]{hyperref}
\journal{Remote Sensing of Environment}

\begin{document}
\begin{frontmatter}
\title{Where Should Physics Enter? Information, Identifiability and Calibration in Probabilistic Sea Surface Temperature Forecasting}

\author[aff1]{Parmitha Vangapandu}
\ead{23bcs092@iiitdwd.ac.in}
\author[aff2]{Jonnakuti Pavan Kumar}
\ead{pavankumar.j@incois.gov.in}
\author[aff3]{Sathwik Narkedimilli}
\ead{sathwik.narkedimilli@u.nus.edu}
\author[aff4]{MSVPJ Sathvik}
\ead{msvpjsathvik@gmail.com}
\author[aff2]{T. M. Balakrishnan Nair}
\ead{director@incois.gov.in}
\author[aff5]{Santanu Bhattacharya\corref{cor1}}
\ead{santanu2@media.mit.edu}
\author[aff6]{Srinivasa Kumar Tummala\fnref{fn1}}
\fntext[fn1]{Mentorship was provided during his time at UNESCO-IOC.}
\cortext[cor1]{Corresponding author}
\affiliation[aff1]{organization={Indian Institute of Information Technology Dharwad},
  city={Dharwad}, country={India}}
\affiliation[aff2]{organization={ODM-ODICT, Indian National Centre for Ocean Information Services (INCOIS)},
  city={Hyderabad}, country={India}}
\affiliation[aff3]{organization={National University of Singapore},
  city={Singapore}, country={Singapore}}
\affiliation[aff4]{organization={University of Birmingham},
  city={Birmingham}, country={United Kingdom}}
\affiliation[aff5]{organization={MIT Media Lab, Massachusetts Institute of Technology},
  city={Cambridge}, state={MA}, country={USA}}
\affiliation[aff6]{organization={Secretary, Ministry of Earth Sciences, Government of India},
  city={New Delhi}, country={India}}

\begin{abstract}
Advection--diffusion priors are a common way to make neural sea surface temperature (SST) forecasts ``physics-informed''. We test where such a prior must enter a heteroscedastic ConvLSTM to help, using daily NOAA OISST v2.1 for the Bay of Bengal (1982--2024) and a synthetic benchmark on the same grid with known currents of controlled strength. (i) On real data, a soft advection--diffusion residual with a learned velocity gives no reliable benefit (lead-15 CRPS $-0.007\,^\circ$C, 95\% interval $[-0.022,+0.008]$); a plain temporal-smoothness penalty matches it, and equivalence within 3\% is not established either. (ii) Learned velocities do not recover the currents, whether placed in the loss, in an explicit transport step, or in a transport step without neural correction: they correlate with the reference no better than a shuffled-date null, even on synthetic data where knowing the currents would improve day-15 forecasts by 16--23\%. (iii) Supplied currents help on synthetic data, but mostly as \emph{information}: giving them to the network as input channels, without transport, recovers 92\% of the transport design's lead-5 gain, and a frozen-current numerical forecast does as well as either. (iv) With GLORYS12 reanalysis currents on real data, the transport design is 6\% and 4\% worse at leads 1 and 5; reanalysis advection explains essentially none of the daily OISST change (correlation 0.02), and frozen-current transport is worse than persistence. (v) The defect that matters is lead-dependent miscalibration shared by every model. One variance multiplier per lead, fitted on validation data, brings 90\% coverage to 0.85--0.91 at all leads and lowers day-1 CRPS by 13--21\%; fine-tuning on 15-day rollouts fixes long-lead coverage but costs short-lead skill. This version withdraws an earlier claim of an 81\% CRPS reduction that came from an uncontrolled comparison.
\end{abstract}

\end{frontmatter}
\sloppy

\section{Introduction}
Sea surface temperature (SST) controls air--sea heat exchange, monsoon rainfall and cyclone intensity, and in the Bay of Bengal it varies on time scales from days (cyclone cold wakes)
to seasons (the monsoon reversal of the East India Coastal Current) \citep{shankar2002monsoon}. Data-driven forecasters built on convolutional recurrent
networks \citep{shi2015convlstm,wang2017predrnn,xiao2019sst} are cheap to run, and a common way to make them ``physics-informed''
\citep{raissi2019pinn,karniadakis2021piml,willard2022integrating,wang2025survey} is an advection--diffusion prior, either as a soft
penalty on the predicted field or as an explicit transport step. \Citet{debezenac2019physical} warp SST with a motion field estimated by a
network and trained without velocity supervision; PhyDNet \citep{leguen2020phydnet} constrains part of a latent space with a PDE. In such designs the velocity is
a latent variable, and whether it is identified, and whether the physics adds anything beyond the information it carries, are rarely tested.

An earlier version of this work reported that a soft residual cut the CRPS of 15-day Bay of Bengal SST forecasts by 81\%.
That comparison was not controlled: the no-physics baseline was trained with mean-squared error, so its variance head was never trained; the inputs were grayscale renderings
of colour-mapped plots rather than SST; the reported ``lead-$h$'' error was a cumulative average; and the upwind stencils of the physics loss were mis-signed. We withdraw those results.

Repeating the experiment properly gives a null result, and the useful questions become \emph{why}, and what would have to change for physics to help. We answer
them with controls that separate three things ``physics-informed'' usually bundles together: a regulariser (the residual), a structure (an explicit transport
step), and information (the velocity field).

The contributions are:
\begin{itemize}
\item A controlled, replicated re-test on real OISST in $^\circ$C (48 models; NLL, $\beta$-NLL, deep-ensemble and statistical baselines; ablations; two re-trained
forecast origins; paired block-bootstrap intervals, equivalence tests and block-length sensitivity). The soft residual gives no reliable benefit and behaves like a smoothness penalty.
\item A synthetic advection benchmark on the real grid with known currents of controlled strength, and controls that isolate information from structure: currents as
input without transport, a frozen-current numerical forecast, a residual with the generator's own operator, and a transport step without a neural correction.
\item Evidence that learned velocities are not recovered in any placement, and that supplied currents help mainly as information.
\item A real-current test with GLORYS12 reanalysis currents \citep{lellouche2021glorys}, plus diagnostics showing why it fails here.
\item Evidence that the dominant fixable defect is lead-dependent miscalibration. A 15-parameter variance correction, a restrictive special case of post-hoc recalibration
\citep{kuleshov2018calibrated}, improves CRPS more than any training-time change we tested, including free-running fine-tuning over the full horizon.
\end{itemize}

\section{Models}\label{sec:method}

\paragraph{Task} Given 45 consecutive daily SST fields on a $56\times48$ grid, predict for each lead $h=1,\dots,15$ days a Gaussian
$\mathcal N(\mu_h(\mathbf x),\sigma^2_h(\mathbf x))$ at every ocean pixel. Forecasts are autoregressive: the predicted mean $x_{h}=\mu_h$
(land set to zero) is the next input.

\paragraph{Backbone} Three gated-attention ConvLSTM cells at full, half and quarter resolution (32 hidden channels each), whose gates receive an extra term
computed from a mix of a fixed 4-neighbour average of the hidden state and four learnable depthwise $3\times3$ stencils (a smoothing operator on hidden features,
not a Laplacian of SST as the first version stated). The scales are upsampled and fused by a $1\times1$ convolution, and $1\times1$ heads output $\mu$, $\log\sigma^2$
and a field $\tilde{\mathbf v}$ mapped to a velocity $\mathbf v=v_{\max}\tanh\tilde{\mathbf v}$ in grid cells per day ($v_{\max}=2$; one cell per day is about 0.3\,m\,s$^{-1}$).

\paragraph{Units and boundaries} All losses are computed on SST standardised by the training-period ocean mean and s.d.\ (one unit $\approx$ s.d.\ of daily SST),
so the residual, conservation and NLL terms share that unit; all reported metrics are converted back to $^\circ$C. Land is zero (the training mean) in inputs, is
excluded from every loss and metric, and outputs are multiplied by the ocean mask. The residual is averaged over interior ocean pixels (all four neighbours ocean),
so its stencils never read land. The transport step samples with border padding at the open edges; near coasts it can sample land values, which is one reason
coastal pixels are reported separately in the diagnostics.

\paragraph{Physics in the loss} The soft residual for lead $h$ is
\begin{equation}
R_h=\mu_h-\big[u_{h-1}-(v_x\,\delta^{\text{up}}_x u_{h-1}+v_y\,\delta^{\text{up}}_y u_{h-1})+\kappa\,\nabla^2_5 u_{h-1}\big],
\end{equation}
with $u_0$ the last observed field, $u_{h-1}=\mathrm{sg}(\mu_{h-1})$ afterwards, first-order upwind differences, the 5-point Laplacian,
$\kappa=0.08$ cells$^2$/day and one-day steps; a unit test checks zero residual for a field translated by $\pm1$ cell/day. The loss is
$\mathcal L_{\text{NLL}}+\lambda_{\text{pde}}\langle R^2\rangle+\lambda_{\text{cons}}(\langle\mu_h\rangle-\langle u_{h-1}\rangle)^2$ with $\lambda_{\text{pde}}=0.75$,
$\lambda_{\text{cons}}=0.05$, $\mathcal L_{\text{NLL}}=\tfrac12\langle (y-\mu)^2/\sigma^2+\log\sigma^2\rangle_{\text{ocean}}$ ($\log\sigma^2\in[-10,5]$).
Neither physics term involves $\sigma^2$. In the \emph{true-velocity} variant (synthetic only) $\mathbf v$ in $R_h$ is the generating current; the
\emph{matched-operator} variant (also with the true $\mathbf v$) further replaces the upwind step by the generator's own semi-Lagrangian step and damping, $R_h=\mu_h-[\mathcal S_{\mathbf v}(u_{h-1})+\kappa\nabla^2\mathcal S_{\mathbf v}(u_{h-1})-\mathcal S_{\mathbf v}(u_{h-1})/\tau]$.

\paragraph{Physics in the model} The mean head instead predicts a correction to an explicit transport step,
\begin{equation}
\mu_h=A_{\mathbf v}(x_{h-1})+c_h,\qquad A_{\mathbf v}(x)=\mathcal S_{\mathbf v}(x)+\kappa\,\nabla^2_5\,\mathcal S_{\mathbf v}(x),
\end{equation}
where $\mathcal S_{\mathbf v}$ is a one-day semi-Lagrangian step (bilinear warp $x(\mathbf r-\mathbf v)$, stable and differentiable in $\mathbf v$, as in
\citealt{debezenac2019physical}) and $c_h$ is the network output. The velocity is \emph{learned} (from the head) or \emph{supplied}: the observed current on
the forecast start date, held fixed over the rollout. A \emph{no-correction} variant sets $c_h\equiv0$, so the network can only act through $\mathbf v$
(and $\sigma$). These arms use $\mathcal L_{\text{NLL}}$ only. Note that the likelihood alone does not identify $\mathbf v$ when $c_h$ is free: for any
$\mathbf v'$, the correction $c'_h=c_h+A_{\mathbf v}(x)-A_{\mathbf v'}(x)$ gives the same $\mu_h$, so nothing in the loss prefers the true current
unless the network finds it easier to represent.

\paragraph{Currents as input (information without structure)} The \emph{velocity-input} arm is NoPhys with the supplied current concatenated to the SST input
as two extra channels at every input step; there is no transport step and no physics loss. Comparing it with the supplied-velocity transport arm separates the
value of the information from the value of the transport structure.

\paragraph{Post-hoc calibration} For a trained model or ensemble we choose one multiplier $s_h$ per lead from a log-spaced grid on $[0.1,10]$ by minimising
CRPS on every third validation window, and forecast with variance $s_h\sigma_h^2$. This is a restrictive special case of post-hoc recalibration: \citet{kuleshov2018calibrated} fit a monotone (isotonic) map of the predictive CDF, which can also correct skew and bias; ours keeps the Gaussian shape and
only rescales it per lead. A \emph{bias+scale} variant first removes a per-lead, per-pixel mean error estimated on the same validation windows. The test period is never used.

\section{Experimental Design}\label{sec:setup}

\paragraph{Real data} NOAA OISST v2.1 daily \citep{huang2021oisst}, $0.25^\circ$, 8.125--21.875\,$^\circ$N, 86.125--97.875\,$^\circ$E, 1 Jan 1982--31 Dec 2024
(15{,}706 days, no gaps; 83\% ocean). Splits are chronological: train to 6 Feb 2012, validation to 20 Jul 2018, test 21 Jul 2018--31 Dec 2024; a window belongs
to the split containing its 15 targets. Rolling-origin runs are re-trained with test years 2008--2011 and 2014--2017.

\paragraph{Reanalysis currents} GLORYS12 \citep{lellouche2021glorys} daily surface currents (0.49\,m, 1993--2024), bilinearly interpolated to the OISST grid;
the eastward component is converted to cells/day with the row's $\cos(\text{latitude})$, the northward with a constant cell length. Coastal OISST cells without
GLORYS currents are set to zero (rms speed 0.82 cells/day). Every arm of this experiment (NoPhys, transport with learned and with GLORYS velocity; 3 seeds each) is trained
on windows from 1993 on, with the main validation and test periods. Learned velocities on the main split are compared with the 117 OSCAR \citep{bonjean2002oscar} 15\,m
fields available to us (2011--2014) using the complex vector correlation \citep{kundu1976veering} and a date-shuffled null. Diagnostics: the correlation and
regression $R^2$ of the GLORYS advective tendency $-\mathbf v\cdot\nabla T$ with the observed next-day change, by season, current-speed tercile, gradient tercile and
coastal vs.\ open pixels, with raw and 2-cell-Gaussian-smoothed currents; a frozen-current transport forecast with no network vs.\ persistence; and the correlation
between the transport increment and the learned correction.

\paragraph{Synthetic benchmark} On the same grid, land mask and dates,
\begin{equation}
T_{t+1}=\mathcal S_{\mathbf v_t}(T_t)+\kappa\nabla^2 T_t-T_t/\tau+\eta_t,
\end{equation}
$\tau=120$ days, $\kappa=0.08$, two semi-Lagrangian half-steps per day, $\eta_t$ white noise smoothed with a 3-cell Gaussian and scaled to 0.1. Currents are
divergence-free eddies $\mathbf v_t=U\nabla^\perp(\psi_t m)$, $\psi_t$ an AR(1) field (20-day memory, 6-cell smoothing), $m$ a smoothed ocean mask that stops flow into land,
with rms speeds 0, 0.5 and 1.0 cells/day. $T$ starts at zero and the first 365 days are discarded as spin-up. All regimes share the noise sequence. References:
perfect physics (generator without noise, true time-varying currents), a \emph{frozen-current numerical} forecast (the model's transport step with the start-date current,
no network), and damped persistence fitted on training data. Arms (3 seeds): NoPhys; loss with learned and true $\mathbf v$; transport with learned and supplied $\mathbf v$
in all regimes; and at 1.0 cells/day additionally the matched-operator residual, the velocity-input arm and the no-correction transport arm.

\paragraph{Training recipe (identical for every arm)} AdamW (lr $2\times10^{-4}$, wd $5\times10^{-4}$), cosine decay over 5{,}000 steps of batch 16,
gradient clipping 1.0, bf16 autocast, random rollout length $L\sim U\{1..5\}$, teacher forcing decaying from 1.0 to 0.2. Every 250 steps we score
256 fixed validation windows with a free 15-day rollout and keep the checkpoint with the lowest validation CRPS. \emph{Free-running fine-tuning} starts
from a trained NoPhys model and trains 750 further steps on fixed 15-day rollouts without teacher forcing, so the variance head sees the horizon it is scored on.

\paragraph{Real-data arms on the main split} NoPhys and loss-Phys (5 seeds); $\beta$-NLL \citep{seitzer2022pitfalls} ($\beta=0.5$); ablations: DiffOnly ($\mathbf v\equiv0$),
Smooth ($\mathbf v\equiv0,\kappa=0$: a plain $\|\mu_h-u_{h-1}\|^2$ penalty), NoCons ($\lambda_{\text{cons}}=0$) (3 seeds each); $\lambda_{\text{pde}}=1.5$ (3), Smooth at weight 1.5 (2),
$\lambda_{\text{pde}}=3$ (1); free-running fine-tuned NoPhys (3); deep ensembles \citep{lakshminarayanan2017ensembles} of the 5 Phys and 5 NoPhys seeds; and climatology,
persistence and damped anomaly persistence fitted on the training period. In total 48 networks are trained on real data and 51 on synthetic data.

\paragraph{Metrics and statistics} Per lead (not cumulative): RMSE, Gaussian(-mixture) CRPS \citep{gneiting2007scoring}, central-interval coverage at 20--90\%,
rms predicted spread, spread/skill ratio and PIT histograms, over ocean pixels of every test window. The \emph{primary endpoints} are lead-15 RMSE and CRPS on the
main test period; other leads, seasons and origins are secondary. Intervals for differences come from a hierarchical bootstrap (2{,}000 replicates) that resamples
seeds within each arm independently and moving blocks of 90 start dates \citep{kunsch1989bootstrap}, with the \emph{same} date blocks for both arms, so the
comparison is paired over dates. We repeat the primary comparison with blocks of 30 and 180 days. ``Supported'' means a 95\% interval excludes zero; for
equivalence we use two one-sided tests (a 90\% interval inside $\pm3\%$ of the NoPhys value, fixed before analysis). For a single ensemble there is no seed
variance to resample; its intervals resample dates only and are conditional on the trained members. Velocity recovery is the mean per-window complex correlation
between learned and reference velocity over interior ocean, against a shuffled-date null (above zero because the statistic is a magnitude), plus the vector RMSE.

\section{Results}\label{sec:results}

\subsection{On real data, the soft physics residual does nothing measurable}\label{sec:real}
\begin{table*}[t]
\caption{Real OISST, main test period (2{,}342 windows, 21 Jul 2018--31 Dec 2024). Per-lead RMSE and CRPS ($^\circ$C), spread/skill ratio, and central-interval
coverage at lead 15. Seeds in parentheses; $\pm$ is the across-seed s.d.\ of lead-15 RMSE. Loss-physics arms use a learned velocity.}
\label{tab:main}
\small
\resizebox{\textwidth}{!}{%
\begin{tabular}{lcccccccc}
\toprule
arm (seeds) & RMSE L1 & L5 & L10 & L15 & CRPS L1 & CRPS L15 & spread/skill L1 / L15 & cov50 / cov90 L15\\
\midrule
climatology & 0.836 & 0.837 & 0.838 & 0.839 & 0.487 & 0.489 & 0.80 / 0.80 & 0.34 / 0.80\\
persistence & \textbf{0.174} & 0.450 & 0.618 & 0.744 & 0.116 & 0.570 & -- & -- \\
damped anomaly persistence & 0.185 & 0.504 & 0.626 & 0.693 & 0.109 & 0.394 & 1.61 / 0.93 & 0.44 / \textbf{0.88}\\
\midrule
NoPhys (5) & 0.178 & 0.442 & 0.572 & 0.668 $\pm$0.017 & 0.114 & 0.386 & 2.07 / 0.54 & 0.32 / 0.66\\
Phys, loss (5) & 0.178 & 0.436 & 0.565 & 0.656 $\pm$0.015 & 0.114 & 0.380 & 2.04 / 0.54 & 0.32 / 0.66\\
$\beta$-NLL (3) & 0.176 & 0.436 & 0.564 & 0.653 $\pm$0.016 & 0.114 & 0.376 & 2.09 / 0.55 & 0.33 / 0.67\\
DiffOnly (3) & 0.178 & 0.435 & 0.565 & 0.660 $\pm$0.009 & 0.114 & 0.382 & 2.05 / 0.54 & 0.32 / 0.66\\
Smooth (3) & 0.178 & 0.434 & 0.563 & 0.656 $\pm$0.014 & 0.113 & 0.380 & 2.03 / 0.54 & 0.31 / 0.66\\
NoCons (3) & 0.177 & 0.434 & 0.564 & 0.658 $\pm$0.020 & 0.114 & 0.380 & 2.06 / 0.54 & 0.32 / 0.67\\
$\lambda_{\text{pde}}=1.5$ (3) & 0.177 & 0.431 & 0.563 & 0.652 $\pm$0.029 & 0.113 & 0.377 & 2.05 / 0.54 & 0.32 / 0.66\\
Smooth, weight 1.5 (2) & 0.177 & 0.432 & 0.568 & 0.667 $\pm$0.014 & 0.114 & 0.388 & 2.07 / 0.53 & 0.31 / 0.65\\
$\lambda_{\text{pde}}=3$ (1) & 0.179 & 0.426 & 0.554 & 0.640 & 0.115 & 0.369 & 2.06 / 0.55 & 0.32 / 0.67\\
\midrule
Ensemble of 5 NoPhys & 0.176 & 0.435 & 0.560 & 0.649 & 0.114 & 0.368 & 2.09 / 0.60 & 0.36 / 0.71\\
Ensemble of 5 Phys & 0.176 & \textbf{0.429} & \textbf{0.554} & \textbf{0.639} & 0.113 & \textbf{0.363} & 2.06 / 0.60 & 0.35 / 0.71\\
\bottomrule
\end{tabular}}
\end{table*}

\begin{table}[t]
\caption{Real OISST: difference from NoPhys in CRPS ($^\circ$C) unless stated; negative favours the first arm. 95\% hierarchical-bootstrap intervals, paired over date blocks;
bold: interval excludes zero. Differences are computed before rounding.}
\label{tab:diff}
\small\setlength{\tabcolsep}{2pt}
\resizebox{\columnwidth}{!}{%
\begin{tabular}{lccc}
\toprule
& lead 1 & lead 5 & lead 15\\
\midrule
Phys (RMSE) & $-0.000$ {\scriptsize[$-.002$,$+.002$]} & $-0.006$ {\scriptsize[$-.013$,$+.001$]} & $-0.012$ {\scriptsize[$-.031$,$+.009$]}\\
Phys & $\mathbf{-0.001}$ {\scriptsize[$-.002$,$-.000$]} & $-0.004$ {\scriptsize[$-.008$,$+.000$]} & $-0.007$ {\scriptsize[$-.022$,$+.008$]}\\
Phys (cov90) & $\mathbf{-0.001}$ {\scriptsize[$-.001$,$-.000$]} & $+0.001$ {\scriptsize[$-.008$,$+.010$]} & $-0.000$ {\scriptsize[$-.019$,$+.018$]}\\
$\beta$-NLL & $-0.000$ {\scriptsize[$-.002$,$+.002$]} & $-0.003$ {\scriptsize[$-.007$,$+.001$]} & $-0.010$ {\scriptsize[$-.025$,$+.006$]}\\
DiffOnly & $-0.001$ {\scriptsize[$-.002$,$+.001$]} & $\mathbf{-0.004}$ {\scriptsize[$-.008$,$-.001$]} & $-0.005$ {\scriptsize[$-.019$,$+.010$]}\\
Smooth & $\mathbf{-0.001}$ {\scriptsize[$-.003$,$-.000$]} & $\mathbf{-0.005}$ {\scriptsize[$-.010$,$-.000$]} & $-0.006$ {\scriptsize[$-.026$,$+.013$]}\\
NoCons & $-0.001$ {\scriptsize[$-.002$,$+.000$]} & $\mathbf{-0.005}$ {\scriptsize[$-.010$,$-.000$]} & $-0.006$ {\scriptsize[$-.027$,$+.014$]}\\
$\lambda_{\text{pde}}=1.5$ & $\mathbf{-0.001}$ {\scriptsize[$-.002$,$-.000$]} & $\mathbf{-0.007}$ {\scriptsize[$-.012$,$-.002$]} & $-0.009$ {\scriptsize[$-.032$,$+.012$]}\\
Phys $-$ Smooth & $+0.000$ {\scriptsize[$-.001$,$+.001$]} & $+0.001$ {\scriptsize[$-.002$,$+.005$]} & $-0.000$ {\scriptsize[$-.018$,$+.019$]}\\
damped pers. & $\mathbf{-0.006}$ {\scriptsize[$-.007$,$-.004$]} & $\mathbf{+0.045}$ {\scriptsize[$+.029$,$+.059$]} & $+0.008$ {\scriptsize[$-.030$,$+.043$]}\\
\bottomrule
\end{tabular}}
\end{table}

With five seeds per arm, Phys and NoPhys are indistinguishable on the primary endpoints (Tables~\ref{tab:main},~\ref{tab:diff}). At lead 15 Phys has 0.012\,$^\circ$C (1.8\%)
lower RMSE and 0.007\,$^\circ$C lower CRPS, both with intervals that include zero; the across-seed s.d.\ of lead-15 RMSE (0.015--0.017\,$^\circ$C) is larger than
the difference. The intervals do not depend on the block length (30, 90 or 180 start dates: RMSE upper bounds $+0.007$ to $+0.009\,^\circ$C, all including zero).
Equivalence is not established either: the 90\% intervals ($[-0.028,+0.005]$ for RMSE, $[-0.019,+0.006]$ for CRPS) leave the pre-set $\pm3\%$ margin (chosen to be close to the across-seed s.d.\ of lead-15 RMSE, 2.5\%) on the
favourable side, so a benefit of up to about 4\% cannot be excluded. We therefore report \emph{no reliable incremental benefit}, not ``no effect''. Secondary
comparisons agree: the intervals that exclude zero at lead 1 belong to effects of 0.001\,$^\circ$C, the four seasons give lead-15 RMSE differences between $-0.025$ and $-0.001\,^\circ$C
with intervals including zero, and re-training at the two earlier origins (test years 2008--2011 and 2014--2017, ``o1'', ``o2'') reverses the sign
(Phys 0.007 and 0.002\,$^\circ$C worse; intervals $[-0.011,+0.024]$, $[-0.021,+0.024]$).

The ablations show what the residual does. Removing advection (DiffOnly), removing advection and diffusion so that only $\|\mu_h-u_{h-1}\|^2$ remains (Smooth),
or removing the conservation term gives the same small lead-5 gain as the full term, and Phys minus Smooth is $-0.000$ $[-0.018,+0.019]$ at lead 15.
A single seed at $\lambda_{\text{pde}}=1.5$ first looked better (lead-15 RMSE 0.621); two further seeds brought the mean to 0.652 $\pm$0.029. On real data the residual
behaves like a temporal-smoothness regulariser on the mean, and we detect no contribution from its advective part. Its learned velocity is not a current: 0.035--0.049\,m\,s$^{-1}$
against 0.21\,m\,s$^{-1}$ for OSCAR, with per-date vector correlation (0.11--0.13) equal to that with date-shuffled OSCAR fields (0.11--0.13); the velocity head of
a DiffOnly model, which receives no gradient, does as well (0.16, null 0.15).

\subsection{Synthetic benchmark: supplied currents help, mostly as information}\label{sec:synth}
Is the null result a property of the Bay of Bengal or of the method? On the synthetic benchmark the currents are known and matter: at day 15 a perfect-physics
forecast is 16--23\% better than the trained networks in the advective regimes (Table~\ref{tab:synth}).

\begin{table}[t]
\caption{Synthetic benchmark: RMSE (synthetic units) per lead, pooled over 3 seeds, and relative lead-15 change from NoPhys. ``Supplied'' $\mathbf v$ = true current
at the forecast start, held fixed. Frozen numerical = the model's transport step with that current and no network. Bold: best network.}
\label{tab:synth}
\small\setlength{\tabcolsep}{3pt}
\resizebox{\columnwidth}{!}{%
\begin{tabular}{llcccc}
\toprule
rms $|\mathbf v|$ & arm & L1 & L5 & L15 & $\Delta$L15\\
\midrule
0 & perfect physics / damped pers. & .100/.100 & .215/.216 & .340/.344 & \\
  & NoPhys & .101 & .215 & .341 & \\
  & loss, learned / true $\mathbf v$ & .101/.101 & .215/.215 & .341/.341 & 0.1\%\\
  & transport, learned $\mathbf v$ & .100 & .214 & .340 & $-$0.1\%\\
\midrule
0.5 & perfect physics / damped pers. & .100/.106 & .210/.249 & .323/.401 & \\
  & frozen numerical (no network) & .100 & .217 & .368 & \\
  & NoPhys & .105 & .242 & .385 & \\
  & loss, learned / true $\mathbf v$ & .105/.105 & .242/.242 & .384/.384 & $-$0.2\%\\
  & transport, learned $\mathbf v$ & .105 & .242 & .385 & $-$0.1\%\\
  & transport, supplied $\mathbf v$ & \textbf{.101} & \textbf{.217} & \textbf{.369} & $-$4.0\%\\
\midrule
1.0 & perfect physics / damped pers. & .100/.116 & .208/.290 & .315/.434 & \\
  & frozen numerical (no network) & .100 & .230 & .401 & \\
  & NoPhys & .113 & .273 & .408 & \\
  & loss, learned / true $\mathbf v$ & .113/.113 & .273/.273 & .405/.406 & $-$0.7/$-$0.5\%\\
  & loss, true $\mathbf v$, matched operator & .113 & .273 & .406 & $-$0.4\%\\
  & transport, learned $\mathbf v$ & .113 & .273 & .406 & $-$0.4\%\\
  & transport, learned $\mathbf v$, no correction & .114 & .282 & .433 & $+$6.2\%\\
  & $\mathbf v$ as input, no transport & .104 & .235 & .402 & $-$1.6\%\\
  & transport, supplied $\mathbf v$ & \textbf{.101} & \textbf{.232} & \textbf{.401} & $-$1.9\%\\
\bottomrule
\end{tabular}}
\end{table}

\begin{figure*}[t]
\centering
\includegraphics[width=\textwidth]{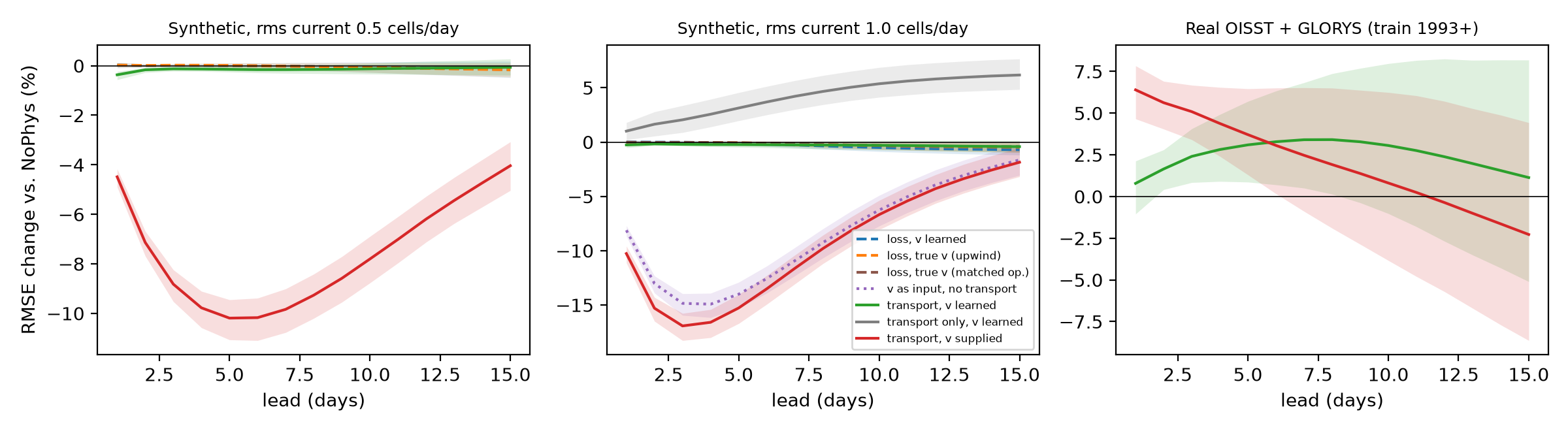}
\caption{RMSE change relative to NoPhys per lead with 95\% hierarchical-bootstrap bands (seeds and paired date blocks). Dashed: physics in the loss; solid: transport
step; dotted: currents as input. Left and centre: synthetic benchmark; right: real OISST with GLORYS12 reanalysis currents (transport arms only).}
\label{fig:where}
\end{figure*}

\emph{Physics in the loss} gains at most 0.7\% at lead 15 even with currents of 1 cell/day, and giving the residual the \emph{true} velocity ($-0.5$\%) or the
generator's own operator and damping ($-0.4$\%, interval $[-0.9,0.0]$\%) does not change this. Operator mismatch is therefore not why the residual fails; the residual
shapes what the network does with information it already has, and the velocity it uses is not available to the network at forecast time.
\emph{Transport with a learned velocity} also gains nothing ($\le0.4$\%). \emph{Transport with the supplied current} helps clearly: RMSE falls by 10\% at lead 5 for
moderate currents and 15\% for strong ones (lead 5, 1.0 cells/day: $-0.042$ $[-0.046,-0.038]$), shrinking with lead as the frozen current decorrelates.

Two controls show that this gain is mostly the information, not the structure (Table~\ref{tab:synth}, Fig.~\ref{fig:where} centre). Feeding the same supplied
current to NoPhys as input channels, with no transport, lowers lead-5 RMSE by $-0.038$ $[-0.042,-0.035]$, 92\% of the transport arm's gain; transport adds
$-0.0035$ $[-0.0045,-0.0026]$ at lead 5 and nothing detectable at lead 15 ($-0.0009$ $[-0.0029,+0.0006]$). And the frozen-current numerical forecast with no network
is as good as the best network at every lead (0.100/0.230/0.401 vs.\ 0.101/0.232/0.401). This benchmark is favourable to the transport arm, because its step
is the generator's own operator; even so, the network adds nothing beyond the numerical step. The learned networks are also miscalibrated here (lead-15 90\% coverage 0.43--0.47 in every arm).

\begin{table}[t]
\caption{Learned velocities are not recovered. Speed (cells/day; OSCAR in m\,s$^{-1}$), per-window vector correlation with the reference vs.\ a shuffled-date null,
and vector RMSE (cells/day; predicting $\mathbf v=0$ gives the reference rms speed); ranges over seeds.}
\label{tab:vel}
\small\setlength{\tabcolsep}{2.5pt}
\resizebox{\columnwidth}{!}{%
\begin{tabular}{llcccc}
\toprule
data & arm & speed (ref.) & corr & null & RMSE\\
\midrule
synth 0.5 & loss, learned & .046--.061 (.445) & .116--.136 & .117--.131 & \\
          & transport, learned & .058--.066 (.445) & .070--.092 & .067--.091 & \\
synth 1.0 & loss, learned & .071--.074 (.890) & .130--.134 & .124--.136 & \\
          & transport, learned & .111--.131 (.890) & .078--.097 & .073--.092 & \\
          & transport, no correction & .177--.235 (.890) & .084--.265 & .080--.132 & .86--.90\\
real      & loss, learned vs OSCAR & .035--.049 m/s (.21) & .106--.134 & .106--.130 & \\
          & transport, learned vs GLORYS & .058--.088 (.817) & .225--.274 & .219--.254 & \\
\bottomrule
\end{tabular}}
\end{table}

\subsection{Learned velocities are not recovered}\label{sec:ident}
In every learned-velocity arm the velocity is 7--26\% of the reference speed, and its correlation with the reference is no better than a shuffled-date null
(Table~\ref{tab:vel}), on synthetic data where the currents matter as well as on real data, in the loss and in the model. Because the likelihood of the transport
model is invariant under $(\mathbf v,c)\mapsto(\mathbf v',\,c+A_{\mathbf v}(x)-A_{\mathbf v'}(x))$ (Sec.~\ref{sec:method}), the collapse is \emph{consistent with}
an ambiguity between transport and neural correction. It is not the whole story: removing the correction, so that the network can change the mean only through
$\mathbf v$, raises the learned speed to 0.18--0.24 cells/day but still does not recover the current (vector correlation 0.27, 0.14 and 0.08 against nulls of 0.13, 0.10 and 0.08;
vector RMSE 0.86--0.90, no better than $\mathbf v=0$), and this arm is 6.2\% worse than NoPhys at lead 15 ($[+4.9,+7.7]$\%). With noise, damping and a frozen
one-day current, a slow, smooth velocity that blurs the field appears to be a better least-squares answer than the true one. The correction does not cancel
the transport where the current is supplied either (Sec.~\ref{sec:glorys}).

\subsection{Reanalysis currents do not help on real data, and diagnostics say why}\label{sec:glorys}
\begin{table}[t]
\caption{Real OISST with physics in the model (trained on 1993--2012 windows, 3 seeds). Difference from NoPhys (1993+) in RMSE ($^\circ$C), 95\% interval;
and diagnostics of the GLORYS12 reanalysis currents (interior ocean, 1993--2024; forecasts on the test period).}
\label{tab:glorys}
\small\setlength{\tabcolsep}{2.5pt}
\resizebox{\columnwidth}{!}{%
\begin{tabular}{lccc}
\toprule
& lead 1 & lead 5 & lead 15\\
\midrule
NoPhys (1993+) RMSE & 0.174 & 0.432 & 0.652\\
transport, learned $\mathbf v$ & $+0.001$ {\scriptsize[$-.002$,$+.004$]} & $\mathbf{+0.013}$ {\scriptsize[$+.004$,$+.025$]} & $+0.007$ {\scriptsize[$-.033$,$+.053$]}\\
transport, GLORYS $\mathbf v$ & $\mathbf{+0.011}$ {\scriptsize[$+.008$,$+.014$]} & $\mathbf{+0.016}$ {\scriptsize[$+.006$,$+.028$]} & $-0.015$ {\scriptsize[$-.058$,$+.029$]}\\
\midrule
persistence RMSE & 0.172 & 0.449 & 0.746\\
frozen GLORYS transport, no network & 0.191 & 0.489 & 0.801\\
\quad same, smoothed currents & 0.188 & 0.486 & 0.803\\
\bottomrule
\end{tabular}}
\end{table}

Driving the transport step with GLORYS12 reanalysis currents makes the real forecast worse at short leads (RMSE 6.4\% higher at lead 1 and 3.7\% at lead 5, intervals
excluding zero; Table~\ref{tab:glorys}, Fig.~\ref{fig:where} right). At lead 15 it is 2.3\% lower in RMSE, with an interval including zero.

The diagnostics explain this without the network. The GLORYS advective tendency $-\mathbf v\cdot\nabla T$ is almost uncorrelated with the observed next-day OISST
change: $r=0.024$ ($R^2=0.001$) overall, and $r\le0.07$ in every season, current-speed and gradient tercile and in coastal and open water, with raw or smoothed currents.
Its rms (0.086\,$^\circ$C/day) is a third of the rms daily change (0.255\,$^\circ$C/day), so the change is dominated by something other than resolved surface advection
of the OISST field, plausibly surface heat fluxes and mixing, and the OISST analysis may not carry the small-scale gradients that reanalysis currents would advect. Consistently,
transporting the start field with frozen GLORYS currents and no network is worse than persistence at every lead (0.801 vs.\ 0.746\,$^\circ$C at lead 15), and smoothing the
currents does not help. In the trained models the learned correction is nearly uncorrelated with the transport increment (0.012--0.036; variance of their sum 1.58--1.71 times
that of the transport alone; rms transport increment 0.065\,$^\circ$C, rms correction 0.050\,$^\circ$C), so the network does not undo the transport, it adds to it. The benchmark's lesson, that a transport prior helps only
through observed currents, needs a further condition on real data: the currents must explain the SST changes being forecast, which here they do not.

\subsection{The defect that matters: lead-dependent miscalibration}\label{sec:recal}
\begin{figure*}[t]
\centering
\includegraphics[width=\textwidth]{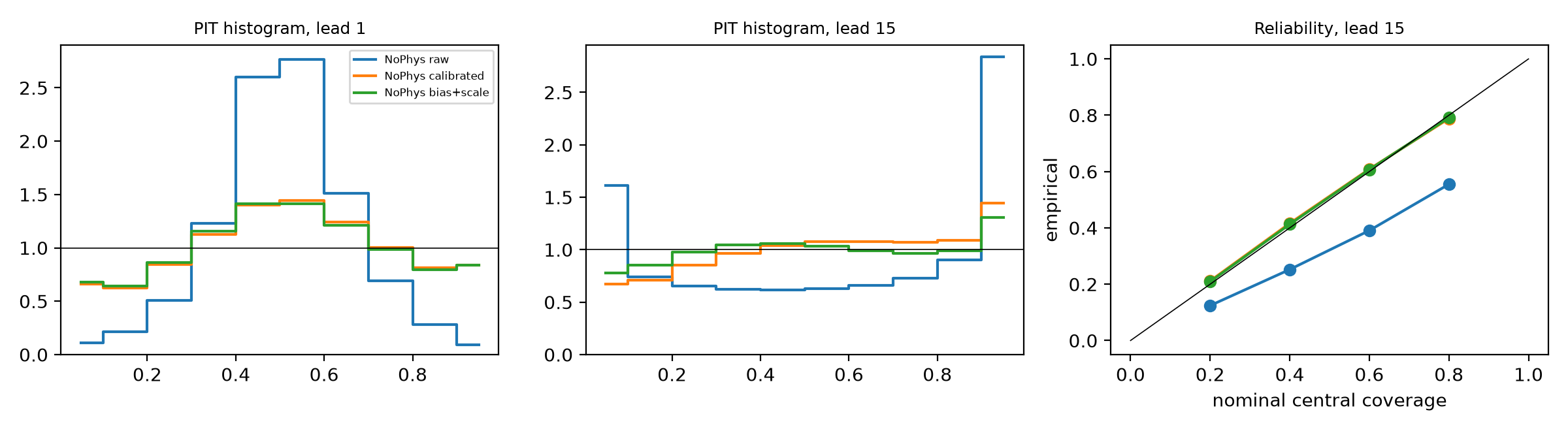}
\caption{Real OISST, 5 NoPhys seeds, test period: PIT histograms at leads 1 and 15 and central-interval reliability at lead 15, before calibration, with a per-lead variance
multiplier, and with per-lead/per-pixel bias removal plus the multiplier.}
\label{fig:cal}
\end{figure*}

\begin{figure*}[t]
\centering
\includegraphics[width=\textwidth]{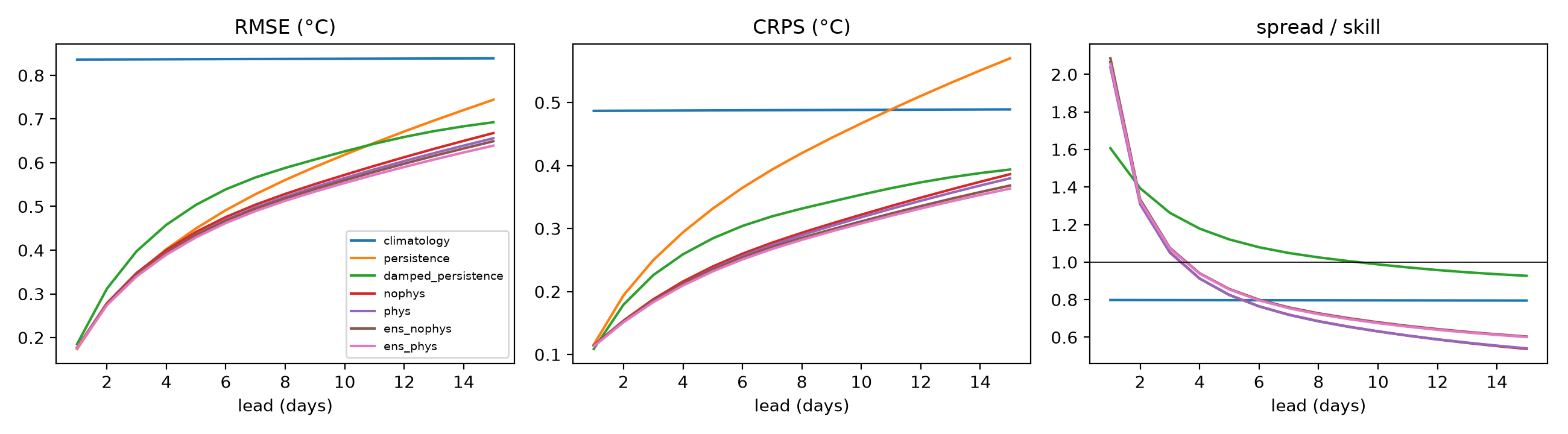}
\caption{Real OISST: per-lead RMSE, CRPS and spread/skill ratio (uncalibrated). Network curves pooled over seeds; persistence has no spread.}
\label{fig:leads}
\end{figure*}

Every network on real data is \emph{under}-confident at short leads and \emph{over}-confident at long ones (Figs.~\ref{fig:cal},~\ref{fig:leads}): spread is about twice the
RMSE at lead 1 and 0.54 times it at lead 15, where 90\% intervals cover 66\% of outcomes and the PIT histogram is U-shaped with a heavier right tail (observations warmer than
forecast). No loss variant changes this (lead-15 spread/skill 0.53--0.55 in every single-model arm). The variance head is trained on short, mostly teacher-forced
rollouts but scored on 15 free-running steps.

\begin{table}[t]
\caption{Post-hoc per-lead variance scaling on real data, re-scored with the training protocol (raw = Table~\ref{tab:main}). CRPS ($^\circ$C) and 90\% coverage
before and after scaling; last column: lead-15 change in CRPS with 95\% interval (for ensembles: over dates only, conditional on the members).}
\label{tab:recal}
\small\setlength{\tabcolsep}{2.5pt}
\resizebox{\columnwidth}{!}{%
\begin{tabular}{lccccc}
\toprule
& \multicolumn{2}{c}{CRPS L1} & \multicolumn{2}{c}{cov90 L1 / L15} & $\Delta$CRPS L15\\
& raw & cal. & raw & cal. & cal.\,$-$\,raw\\
\midrule
NoPhys (5) & 0.114 & \textbf{0.091} & 0.99 / 0.66 & 0.91 / 0.88 & $-0.020$ {\scriptsize[$-.038$,$-.005$]}\\
Phys (5) & 0.114 & \textbf{0.091} & 0.99 / 0.66 & 0.91 / 0.88 & $-0.019$ {\scriptsize[$-.033$,$-.007$]}\\
Ens.\ NoPhys & 0.114 & \textbf{0.090} & 0.99 / 0.71 & 0.91 / 0.87 & $-0.012$ {\scriptsize[$-.018$,$-.008$]}\\
Ens.\ Phys & 0.113 & \textbf{0.090} & 0.99 / 0.71 & 0.91 / 0.87 & $-0.013$ {\scriptsize[$-.018$,$-.008$]}\\
o1 NoPhys (2) & 0.127 & \textbf{0.109} & 0.98 / 0.67 & 0.90 / 0.91 & $-0.017$ {\scriptsize[$-.027$,$-.008$]}\\
o1 Phys (2) & 0.129 & \textbf{0.109} & 0.98 / 0.68 & 0.90 / 0.91 & $-0.017$ {\scriptsize[$-.027$,$-.008$]}\\
o2 NoPhys (2) & 0.127 & \textbf{0.110} & 0.98 / 0.68 & 0.85 / 0.88 & $-0.016$ {\scriptsize[$-.037$,$+.003$]}\\
o2 Phys (2) & 0.127 & \textbf{0.110} & 0.98 / 0.68 & 0.85 / 0.89 & $-0.016$ {\scriptsize[$-.030$,$-.004$]}\\
\bottomrule
\end{tabular}}
\end{table}

A per-lead variance multiplier fitted on validation data removes most of this (Table~\ref{tab:recal}). The multipliers are nearly identical across seeds, arms and origins
(about 0.2 at lead 1, 2.5--3.4 at lead 15). Calibration lowers lead-1 CRPS by 13--21\% in every case and lead-15 CRPS by 3--5\% (interval excluding zero in 7 of 8 cases),
leaves lead 5 unchanged, and brings 90\% coverage to 0.85--0.91 at every lead. Reliability holds across levels: at lead 15 the calibrated central 20/40/60/80/90\% intervals
cover 21/42/61/79/88\% of outcomes (raw: 12/25/39/55/66\%), with rms predicted spread rising from 0.36 to 0.62\,$^\circ$C at lead 15 and falling from 0.37 to 0.17\,$^\circ$C at lead 1.
It composes with ensembling: the calibrated Phys ensemble has the lowest CRPS of any forecast at leads 1, 5 and 15 (0.090, 0.232, 0.350). After calibration Phys and NoPhys
remain indistinguishable (lead-15 CRPS $-0.006$, $[-0.018,+0.007]$).

\begin{table}[t]
\caption{Calibration remedies, NoPhys seeds 0--2, test period: CRPS, RMSE ($^\circ$C) and 90\% coverage. Fine-tuned = 750 extra steps on 15-day free-running rollouts.}
\label{tab:remedy}
\small\setlength{\tabcolsep}{2.5pt}
\resizebox{\columnwidth}{!}{%
\begin{tabular}{lccccc}
\toprule
& \multicolumn{3}{c}{CRPS L1 / L5 / L15} & RMSE L1 / L15 & cov90 L1 / L15\\
\midrule
raw & .114 & .239 & .382 & .177 / .661 & .99 / .66\\
variance scale & .090 & .238 & .363 & .177 / .661 & .91 / .88\\
bias + scale & .090 & .235 & \textbf{.351} & .176 / .640 & .91 / .88\\
free-running fine-tuned (raw) & .172 & .261 & .375 & .214 / .683 & 1.00 / .88\\
\bottomrule
\end{tabular}}
\end{table}

Two alternatives target the same defect (Table~\ref{tab:remedy}). Fine-tuning on full 15-day free-running rollouts teaches the variance head the long horizon
(lead-15 coverage 0.66$\to$0.88) and slightly lowers lead-15 CRPS, but makes short leads much worse (lead-1 CRPS 0.114$\to$0.172, RMSE 0.177$\to$0.214, coverage 1.00):
the mean and variance trained for day 15 no longer fit day 1. Post-hoc scaling of that model was not possible within our grid because the optimal lead-1 multiplier
is below 0.1. Removing a per-lead, per-pixel bias before scaling gives a further 3\% at lead 15 (0.351 vs.\ 0.363; five seeds: 0.357 vs.\ 0.367). These remedy
comparisons are point estimates over three seeds without intervals. A global scale cannot vary in space: for one Phys model it leaves the strongest-gradient third of pixels
(90\% coverage 0.84) and the coastal strip (0.82) over-confident. A separate scale per group restores coverage there but changes pooled CRPS by only 0.2\%; we did not
test richer spatial or state-dependent variance models.

\paragraph{Baselines and ensembles} The networks beat climatology at every lead, persistence in RMSE from lead 6, and damped anomaly persistence in CRPS at leads 2--10
(lead 5: 0.240 vs.\ 0.284\,$^\circ$C). By lead 15 the uncalibrated networks are within noise of damped persistence, which is better calibrated (spread/skill 0.93).
A five-seed deep ensemble lowers lead-15 CRPS by 4.7\% and raises 90\% coverage from 0.66 to 0.71. The Phys ensemble is 0.005\,$^\circ$C better than the NoPhys
ensemble at lead 15; with one ensemble per arm this has no seed-level interval and we do not count it as evidence for the physics term.

\section{Discussion}
\paragraph{Where physics should enter} The controls separate the regulariser, the structure and the information. On data where currents matter, the regulariser
adds almost nothing, even with the exact velocity and the generator's own operator inside it; a transport step adds nothing when the velocity is learned, because the
learned velocity is not recovered; and when the velocity is supplied, most of the gain comes from the currents themselves, which a plain network uses about as well when
given them as input. The transport structure adds a small, significant short-lead gain on top, on a benchmark built from that very operator. For latent-velocity designs
in the spirit of \citet{debezenac2019physical} our evidence suggests two checks before claiming physical content: compare the learned velocity with an independent current
product against a shuffled null, and compare the transport model with a network that receives the same inputs without the transport.

\paragraph{When the conclusion applies} The real-data failure has a measurable cause here: reanalysis advection explains essentially none of the daily OISST change, and
pure transport is worse than persistence. Any physics prior built on surface advection should be preceded by this check. Regions or products where it passes, such as
western boundary currents or higher-resolution SST, are where a transport prior could help; we did not test them.

\paragraph{What did help} A 15-parameter post-hoc variance correction improved probabilistic skill more than any training-time change we tried, including direct
free-running training over the horizon. Training-time remedies should be compared against this baseline, not against uncalibrated NLL.

\paragraph{Limitations and what we did not do} One region, one resolution and one backbone; 3--5 seeds per arm, so lead-15 effects of a few per cent on real data cannot
be excluded; hyperparameters inherited from the first version; synthetic dynamics linear in the tracer with idealised eddies, generated with the model's own transport
operator; supplied currents frozen at the start date; pixel-wise metrics only. GLORYS is a model reanalysis interpolated from 1/12$^\circ$ to 0.25$^\circ$. The velocity-input
control and the no-correction arm were run on synthetic data only, the real-data transport experiment has 3 seeds, and the remedy comparisons in Table~\ref{tab:remedy} have no
intervals. Within our compute budget we did not run CRPS fine-tuning, a velocity-supervised arm, alternative velocity initialisations, a second backbone or region, or a
validation split separate from the one used to select checkpoints and fit the calibration (the test period is never used for either). Conclusions rest on the 3--5-seed arms;
1--2-seed arms are reported for completeness. The bootstrap treats fitted calibration multipliers as fixed, so it does not include their estimation variance. On the synthetic benchmark the operator a network must learn is a simple linear warp; that input information sufficed there need not hold for dynamics a network finds harder to learn.

\section{Conclusion}\label{sec:conclusion}
We asked where an advection--diffusion prior has to enter a probabilistic ConvLSTM to improve sea surface temperature forecasts, and separated three things
that ``physics-informed'' usually bundles together: a regulariser, a transport structure and the velocity information. Three findings follow.

\emph{Physics as a regulariser does not help.} On 43 years of daily Bay of Bengal OISST, a soft advection--diffusion residual changes lead-15 CRPS by
$-0.007\,^\circ$C (95\% interval $[-0.022,+0.008]$). It is matched by a plain temporal-smoothness penalty, and its learned velocity correlates with OSCAR
currents no better than a date-shuffled null.

\emph{A learned velocity is not identified, and supplied currents help mainly as information.} The learned velocity is not recovered in the loss, in an
explicit transport step, or in a transport step without a neural correction, even on a synthetic benchmark where a forecast that knew the currents would be
16--23\% better at day 15. Physics helps only when the currents are supplied, and then mostly through the information they carry: giving them to the network
as input channels, with no transport, recovers 92\% of the transport design's lead-5 gain. On real data, GLORYS12 currents do not help, because reanalysis
advection explains essentially none of the daily OISST change (correlation 0.02).

\emph{The defect that matters is calibration.} Every model is miscalibrated in a lead-dependent way. One variance multiplier per lead, fitted on validation
data, brings 90\% coverage to 0.85--0.91 at every lead and lowers day-1 CRPS by 13--21\%, more than any training-time change we tested.

These results translate into three checks for physics-informed forecasting. Before adding a transport prior, test whether the advective tendency of an
independent current product explains the observed day-to-day change. Compare any physics model with a network that receives the same inputs without the
physics, and compare learned velocities with independent currents against a shuffled null before attributing physical meaning to them. Report calibration
per lead, and use post-hoc recalibration, not uncalibrated likelihood training, as the baseline for training-time uncertainty methods. The natural next tests
are regions and products where surface advection does explain the SST tendency, such as western boundary currents or higher-resolution SST, and designs in
which the velocity is supervised or assimilated rather than learned without supervision.

\section*{Declaration of competing interest}
The authors declare that they have no known competing financial interests or personal relationships that could have appeared to influence the work reported in this paper.

\section*{Data availability}
All data are public: NOAA OISST v2.1, GLORYS12 (Copernicus Marine Service) and OSCAR. Data download (\texttt{get\_data.py}, \texttt{get\_currents.py}), the
synthetic generator (\texttt{make\_synth.py}), training and evaluation (\texttt{exp.py}, with unit tests in \texttt{test\_exp.py}), analysis
(\texttt{analyze.py}, \texttt{analyze\_synth.py}, \texttt{glorys\_diag.py}, \texttt{fig\_where.py}) and the run log are provided as supplementary material;
every number in the paper is produced by these scripts from saved per-window metrics. Models were trained on rented RTX 3090/4090 GPUs (about 80 minutes
per run). This version replaces an earlier one whose central claims are withdrawn for the reasons given in the introduction.

\section*{Declaration of generative AI and AI-assisted technologies in the writing process}
During the preparation of this work the authors used Claude (Anthropic) and Gemini (Google) for coding assistance, experiment orchestration and draft review.
After using these tools, the authors reviewed and edited the content as needed, checked all results against the saved outputs, and take full responsibility
for the content of the publication.

\bibliographystyle{elsarticle-harv}
\bibliography{refs}
\end{document}